%% file: egpaper_for_review.tex
\documentclass[10pt,twocolumn,letterpaper]{article}

\usepackage{iccv}
\usepackage{times}
\usepackage{epsfig}
\usepackage{graphicx}
\usepackage{amsmath}
\usepackage{amssymb}

\usepackage{enumitem}
\usepackage{bm}
\usepackage{booktabs}
\usepackage{multirow}

\usepackage[pagebackref=true,breaklinks=true,letterpaper=true,colorlinks,bookmarks=false]{hyperref}

\iccvfinalcopy % *** Uncomment this line for the final submission

\def\iccvPaperID{4270} % *** Enter the ICCV Paper ID here

\ificcvfinal\fi
\begin{document}

%%%%%%%%% TITLE
\title{Learning Online Scale Transformation for Talking Head Video Generation}

\author{Fa-Ting Hong\\
CSE, HKUST\\
{\tt\small fhongac@connect.ust.hk}
% For a paper whose authors are all at the same institution,
% omit the following lines up until the closing ``}''.
% Additional authors and addresses can be added with ``\and'',
% just like the second author.
% To save space, use either the email address or home page, not both
\and
Dan Xu\\
CES, HKUST\\
{\tt\small fhongac@connect.ust.hk}
}

\maketitle
% Remove page # from the first page of camera-ready.
\ificcvfinal\thispagestyle{empty}\fi

%%%%%%%%% ABSTRACT
\begin{abstract}
   % One-shot talking head video generation takes a source image and a driving video as input to generate a synthetic video in which the person in the source image animates the facial movements of the driving video. During practice, the movement of the face in the driving video can cause the scale of the driving face to varying, while the scales of the source and driving images are typically different in the cross-identity facial reenactment. Most current methods aim to find the best-aligned frame in the driving video with the source image to use as the anchor frame for relative motion transfer, but the imprecise alignment of the anchor frame with the source image can lead to suboptimal results. 
   One-shot talking head video generation uses a source image and driving video to create a synthetic video where the source person's facial movements imitate those of the driving video. However, differences in scale between the source and driving images remain a challenge for face reenactment. Existing methods attempt to locate a frame in the driving video that aligns best with the source image, but imprecise alignment can result in suboptimal outcomes.
   To this end, we introduce a scale transformation module that can automatically adjust the scale of the driving image to fit that of the source image, by using the information of scale difference maintained in the detected keypoints of the source image and the driving frame. Furthermore, to keep perceiving the scale information of faces during the generation process, we incorporate the scale information learned from the scale transformation module into each layer of the generation process to produce a final result with an accurate scale. Our method can perform accurate motion transfer between the two images without any anchor frame, achieved through the contributions of the proposed online scale transformation facial reenactment network. Extensive experiments have demonstrated that our proposed method adjusts the scale of the driving face automatically according to the source face, and generates high-quality faces with an accurate scale in the cross-identity facial reenactment. 
\end{abstract}

%%%%%%%%% BODY TEXT
\vspace{-5pt}
\section{Introduction}
\vspace{-5pt}
\begin{figure}[ht]
  \centering
    \includegraphics[width=1\linewidth]{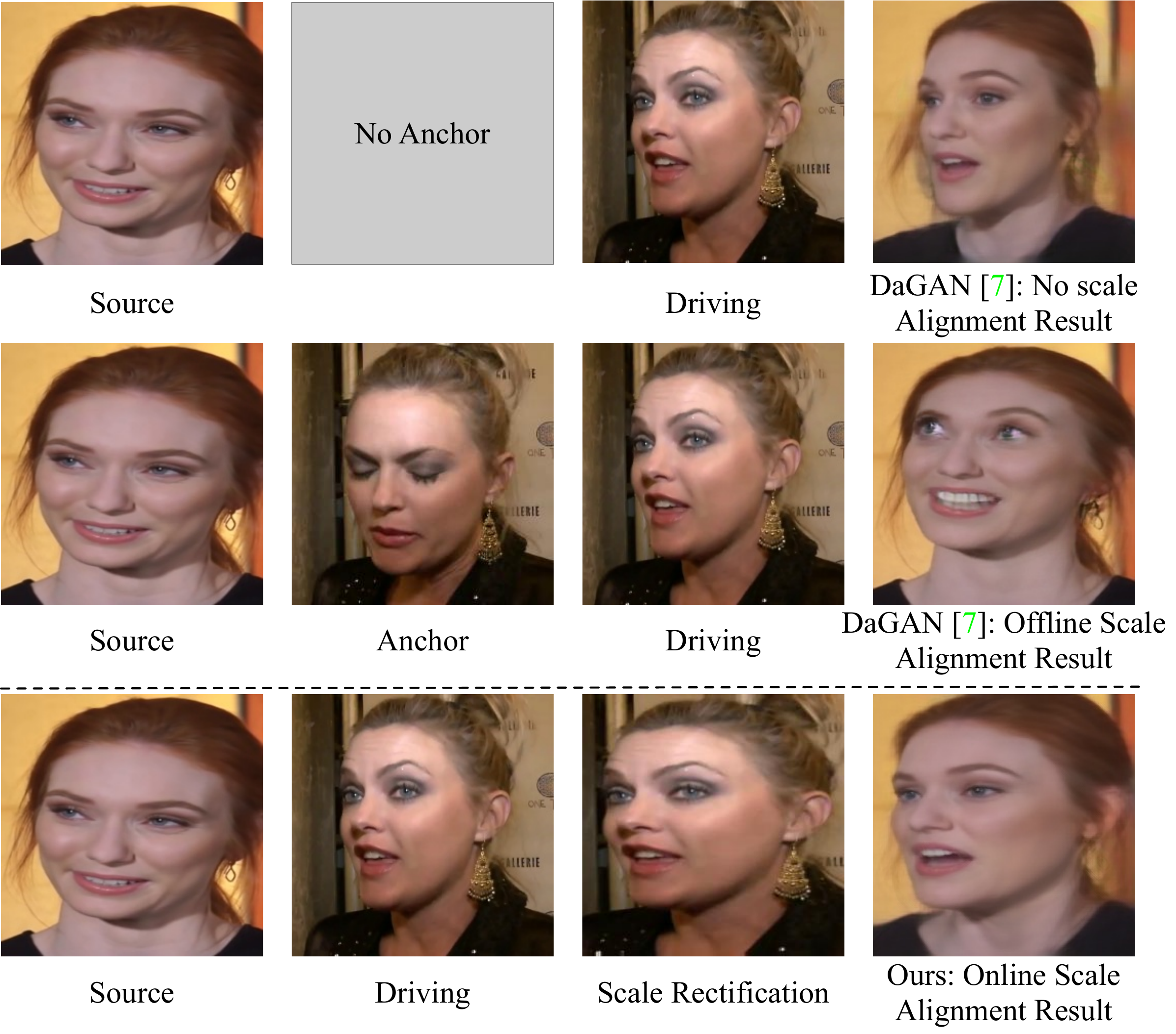}
    \vspace{-15pt}
    \caption{
    % The methods proposed in~\cite{hong2022depth, siarohin2019first} are sensitive to the scale of the driving face. If scale alignment is not performed between the source and driving images, the quality of the results can be severely impacted by the scale of the driving face (as shown in the first row). Therefore, these methods typically search offline through the frames of the driving video to find the best anchor frame that can be aligned with the source image. However, if the anchor frame is not perfectly aligned with the source face, the resulting output will be suboptimal (see the second row of Figure~\ref{fig:firstpage}). In our method, we aim to transform the scale of the driving face to the source face first using the proposed scale transformation module (see ``Scale Rectification'' in the third row). Thus, we do not need to search for a perfectly aligned anchor frame and can produce good quality animated results (see the result of ``Ours'' in the last row). 
    Methods in~\cite{hong2022depth, siarohin2019first} are sensitive to driving face scale (the first row) and rely on offline searching for an anchor frame. However, imperfect alignment can result in suboptimal outputs (the second row). Our method adjusts the driving facial scale to match the source face, eliminating the need for anchor frame searching and producing high-quality animated results (the last row).
    }
    \vspace{-10pt}
    \label{fig:firstpage}    
\end{figure}

%------------------------------------------------------------------------
In this work, we study the task of generating a video of a human head using one source image of the person and a driving video, possibly derived from another person. The person in the generated video performs the motions provided by the driving video. One-shot talking head video generation has extensive applications in the real world, including movie production, photography, and virtual avatar. Rapid progress has been made in talking head video generation in recent years. Existing image-driven methods can be divided into two categories: model-based and model-free methods. Model-based methods~\cite{yao2020mesh,zhao2021sparse,ren2021pirenderer,yin2022styleheat} use third-party pre-trained models (\eg, 3DMM model~\cite{blanz1999morphable,yao2021learning} and landmark model~\cite{wood20223d}) to extract motion representations from driving images. In contrast to model-based methods, model-free methods~\cite{siarohin2019first,wang2021one,hong2022depth} attempt to learn the motion representation between two face images by detecting the keypoints of the human face.
% For example, StyleHeat~\cite{yin2022styleheat} uses a pretrained 3DMM~\cite{blanz1999morphable} to extract motion parameters from driving videos and then injects them into CNN layers via adaptive instance normalization (AdaIN~\cite{huang2017arbitrary}) to generate a motion field. 

 However, those model-based methods suffer from error accumulation due to the inaccuracy of the pre-trained model. 
For model-free methods, the facial scale presented by the keypoints of the driving face introduces identity noise that would affect the final generation result (see the first row of Figure~\ref{fig:firstpage}).
To eliminate the negative impact of facial scale during facial motion transfer, existing model-free methods have to find a frame in the driving video that best aligns with the source image, and use it as an anchor frame for relative motion transfer. Unfortunately, imprecise alignment of the anchor frame with the source image can lead to suboptimal results when the driving video lacks a frame to align accurately with the source image (see the second row of Figure~\ref{fig:firstpage}). 
 % While TPSM~\cite{zhao2022thin} and MRAA~\cite{siarohin2021motion} attempt to address this limitation by applying a transformation affine on keypoints/regions and training an additional shape and pose encoder during the testing stage, these methods may fail when the scales of the source image and driving frame differ significantly (see Figure~\ref{fig:cross-id}). Additionally, the transformation affine applied to the detected keypoints/regions in these methods can potentially modify the facial expression and pose information maintained in the keypoints/regions. 
 To this end, we propose an approach that rectifies the scale of the driving face to match that of the source face online. By so doing, we can facilitate more accurate motion transfer between the two images without relying on a best-aligned anchor frame, to produce the final generation result with an accurate scale (see the third row of Figure.~\ref{fig:firstpage}).

 To enable the model to rectify the scale of the driving face to fit the source image, the model should not only be aware of the source scale but also handle a driving face of any scale. Therefore, we utilize an expression-preserved augmentation to remove the original scale information in the driving face, while maintaining the head pose and the facial expression to produce training pairs with different facial scales in the training stage. Our method consists of two steps to create a high-quality result with an accurate scale: (i)~Scale transformation. We introduce a scale transformation module to align the scale of the driving face with the source face online by utilising the scale information embedded in the detected keypoints of images. (ii)~Scale embedding. The scale information can progressively degrade as it passes through multiple layers during the generation. To keep perceiving the scale information of faces, we embed the latent scale code that is learned from the proposed scale transformation module, into each layer of the generation process to produce the final result with a more accurate scale. 
All the above-introduced contributions compose a \textbf{O}nline \textbf{S}cale \textbf{T}ransformation facial reenactment \textbf{Net}work for talking head video generation, and we coin it as \textbf{OSTNet}. 
% for simplicity. 
Extensive experiments are performed to qualitatively and quantitatively evaluate our proposed OSTNet model on two talking head generation datasets, \ie, VoxCeleb1~\cite{nagrani2017voxceleb} and HDTF~\cite{wu2018reenactgan}. The experimental results illustrate that our proposed scale transformation module can automatically and correctly adjust the scale of the driving faces to align with that of the source face. Equipped with the proposed scale transformation module and the multi-layer scale embedding in the generation process, our OSTNet can generate high-quality video results with any facial scale, without using a best-aligned anchor frame, which is significantly different from existing methods. 

In summary, our contribution of this paper is threefold:
\begin{itemize}
    \vspace{-5pt}
    \item To the best of our knowledge, we are the first to address the problem of facial scale alignment online on talking head video generation. The model can precisely transfer the facial motion between two faces without an intermediate anchor frame by aligning the driving facial scale with the source facial scale.
    \vspace{-5pt}
    \item We propose a novel online scale transformation facial reenactment network (OSTNet) for talking head video generation trained in an end-to-end manner, which employs a scale transformation module to effectively adjust the scale of the driving face online while embedding the scale information into the generation process to produce identity-preserve results.
    \vspace{-5pt}
    \item We conduct extensive experiments to show that our proposed scale transformation method can automatically adjust the scale of the driving face to match that of the source face. The experimental results show that our method outperforms the existing methods on all evaluation metrics and significantly improves on different scale face animation.
    \end{itemize}

\begin{figure*}[t]
  \centering
    \includegraphics[width=1\linewidth]{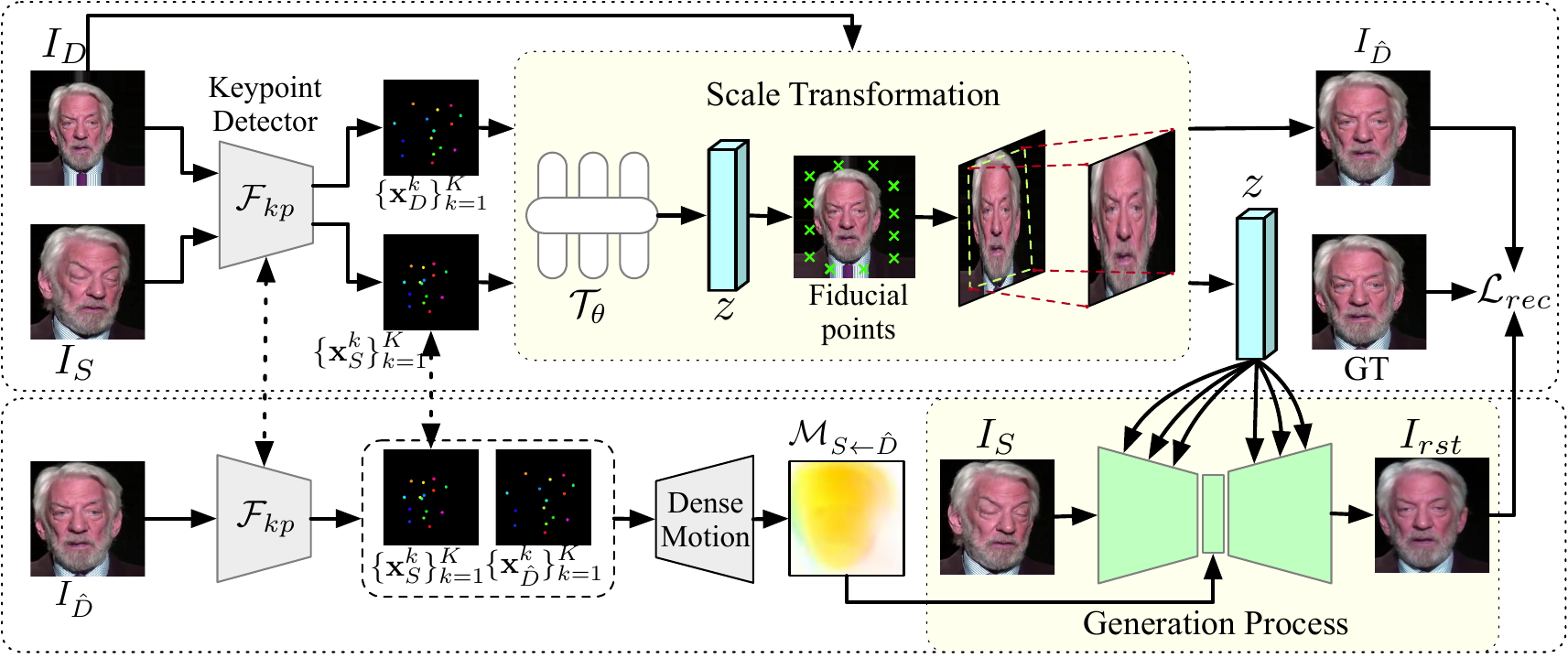}
    \vspace{-10pt}
    \caption{The framework of our OSTNet. We proposed a scale transformation module to align the scale of the source face and the driving face to produce a scale rectified image $I_{\hat{D}}$ using the scale information extracted from both keypoints of the source image $I_S$ and the driving image $I_{D}$. We embed the scale information $z$ learned from the scale transformation module into the generation process to produce the result with a more accurate scale. In this way, our method can automatically adjust the scale of driving face online and produce results with an accurate scale. We utilize the reconstruction loss $\mathcal{L}_{rec}$ to facilitate learning our scale transformation module and the whole network.}
    \vspace{-5pt}
    \label{fig:framework}    
\end{figure*}
\vspace{-5pt}
\section{Related Works}
\vspace{-5pt}
%------------------------------------------------------------------------
\noindent\textbf{Talking head video generation.} Talking head video generation~\cite{tang2022explicitly,drobyshev2022megaportraits,yao2020mesh,ren2021pirenderer,siarohin2019animating,wang2021one,zeng2022fnevr,shen2023difftalk,gururani2022spacex,ji2022eamm} has gained increasing attention recently due to its potential application value in various industries. Typically, there are two categories for generating talking head videos based on driving modalities: image-driven and audio-driven. Image-driven talking head video generation~\cite{siarohin2019first,hong2022depth,yin2022styleheat,drobyshev2022megaportraits,yao2020mesh,ren2021pirenderer,siarohin2019animating,wang2021one} uses a source image and a driving video as input to synthesize a realistic-looking video. Specifically, FOMM~\cite{siarohin2019first}, face-vid2vid~\cite{wang2021one}, and DaGAN~\cite{hong2022depth} utilize a first-order approximation to calculate the facial motion between two faces by the detected facial keypoints. Different from these methods that they learned the facial motion in a self-supervised manner, other methods~\cite{yin2022styleheat, ren2021pirenderer, yao2020mesh} adopt a third-party pre-trained model (\eg, 3DMM~\cite{blanz1999morphable}) to extract the motion parameters, which will be utilized to construct the motion flow between the source image and the driving frame. The audio-driven methods~\cite{zhou2021pose, wu2021imitating,zhou2019talking,prajwal2020lip} take the audio as the driven modality and then map the audio information into the motion flow of the lip. DAVS~\cite{zhou2019talking} disentangled the joint audio-visual embedding space from the person-ID space through adversarial learning, while the PC-AVS~\cite{zhou2021pose} adopt the contrastive learning protocol to seek the synchronization between audio and visual features.

We aim to generate talking head videos without using a pre-trained model and solely relying on images. Unlike prior research, we design a scale transformation module to adjust the driving frame's scale to match the source image. The proposed approach allows us to create scale-sensitive video without the need of a perfectly aligned anchor frame.

\noindent\textbf{Face/Scale Alignment.} Face alignment~\cite{zhu2015face, zhang2014coarse,zhang2014facial} has been a well-studied task for a long time and is often considered an intermediate step in commonly used face analysis pipelines. In this work, we focus on aligning the scale of two faces for talking head generation purposes. Typically, the scales differ between the source and driving faces, while they have different identities. 
% The model-based methods~\cite{yao2020mesh,zhao2021sparse,ren2021pirenderer,yin2022styleheat,doukas2021headgan} can use disentangled expression code provided by pre-trained model to avoid this problem, but they inevitably meet the error accumulation caused by the inaccuracy of the pre-trained model. In contrast, 
Model-free methods~\cite{hong2022depth,siarohin2021motion,zhao2022thin,siarohin2019first,wang2021one} can learn accurate motion between two faces.
% Given a face, these methods detect a set of keypoint to represent the expression and head pose. But the distribution of the keypoints also maintains the scale of the corresponding face, which can introduce noise into the generation process. 
To address the problem of scale difference between the source and driving image, Taylor approximation methods ~\cite{siarohin2019first,hong2022depth,wang2021one} search the best-aligned frame offline in the driving video with the source image to use as the anchor frame for relative motion transfer. Furthermore, both methods~\cite{siarohin2021motion, zhao2022thin} train an additional shape and pose encoder offline in the second stage to recover the distribution of keypoints that have been deformed by a random transformation.

Unlike previous works, our OSTNet synthesizes a high-quality face with accurate scale by end-to-end online joint learning of scale transformation and generation. We introduce a scale transformation module to adjust the scale of the driving face to align with the source face in an online mode. In this way, we can train our method end-to-end and do not need to find a best-aligned anchor frame.

% \noindent\textbf{Image transformation.} Spatial Transformer Networks (STN~\cite{jaderberg2015spatial}) is a (sub-)differentiable module that performs spatial transformations on images or feature maps. It has gained significant attention in the deep learning community because it reduces geometric variations in the data. IC-STNs~\cite{lin2017inverse} theoretically connect STN with the Lucas \& Kanade algorithm~\cite{lin2016conditional} to show how multiple spatial transformations within an alignment framework can eliminate geometric variations more efficiently. Furthermore, RNN-SPN~\cite{sonderby2015recurrent} integrates STN into a recurrent neural network to classify digits in cluttered MNIST sequences.

% In this work, we introduce a scale-conditioned spatial transformer module in talking head video generation to adjust the scale of the driving image by the scale information maintained in the detected keypoints.

\vspace{-5pt}
\section{Methodology}
\vspace{-5pt}
The inconsistency of facial scale between the source face and the driving face is an inevitable problem in this task. We propose an online scale transformation facial reenactment network (OSTNet) to automatically adjust the scale of the driving face, so that our method can generate precise results without finding a best-aligned anchor frame.

% 1) scale removal. Given a source image $I_S$ and a driving image $I_D$, we first apply an expression-preservedaugmentation (\ref{sec:transform}) on $I_D$ to produce an augmented image $I_{D^\prime}$, which contains the same expression as the driving image $I_D$ but with the different scale. After that, The keypoint detector $\mathcal{F}_{kp}$ takes both $I_{D^\prime}$ and $I_S$ as input and produces two sets of keypoints $\{\mathbf{x}_{D^\prime}^k\}_{k=1}^K$ and $\{\mathbf{x}_{S}^k\}_{k=1}^K$, repsectively. 
\vspace{-3pt}
\subsection{Overview}
\vspace{-3pt}
As illustrated in Figure~\ref{fig:framework}, our method can be divided into two steps: 1) Scale transformation. We introduce a scale transformation module (Sec.~\ref{sec:transform}) to align the scale of the driving face $I_D$ with that of the source image $I_{S}$. Specifically, the scale transformation module embeds the scale information using the keypoints $\{\mathbf{x}_{D}^k\}_{k=1}^K$ and $\{\mathbf{x}_{S}^k\}_{k=1}^K$ to predict a set of fiducial points $\mathbf{C} = \{\mathbf{c}_i\}_{i=1}^T$ (The $T$ is the pre-defined number of the fiducial points). The predicted fiducial points are then fed into the grid generator to produce a scale deformation map $\mathcal{M}_{tran}$, which will be utilized to warp the driving image $I_{D}$ to produce a scale rectified image $I_{\hat{D}}$. 2) Scale embedding. To make the network aware of the scale of the source image during face generation, we aggregate the latent scale code $z$ learned from the scale transformation module into each layer of the generation process (Sec.~\ref{sec:inject}). In this way, our method can further constrain the scale on the face generation.

In the training stage, we apply an expression-preserved augmentation on driving images to produce training pairs with different scales so that our method can handle the driving face of any scale. We take the augmented image as the driving face and feed it into our framework with the source image to produce the final result. 
% Our method will automatically rectify the scale of the driving face.

\vspace{-3pt}
\subsection{Scale-aware spatial transformation} 
\vspace{-3pt}
In real applications, the identities of the source image and the driving video are usually different. Thus, facial scale inconsistency arises from variations in the identities of faces, while the movement of the head in the video could also cause the scale to vary. To eliminate the identity noise introduced by the driving facial scale in the final result, we aim to design a model capable of adjusting the scale of the inputted driving face to match that of the source face, thereby ensuring that the final result maintains the identity of the source face. To this end, we propose a scale transformation module to rectify the facial scale online.

\begin{figure}[t]
  \centering
    \includegraphics[width=1\linewidth]{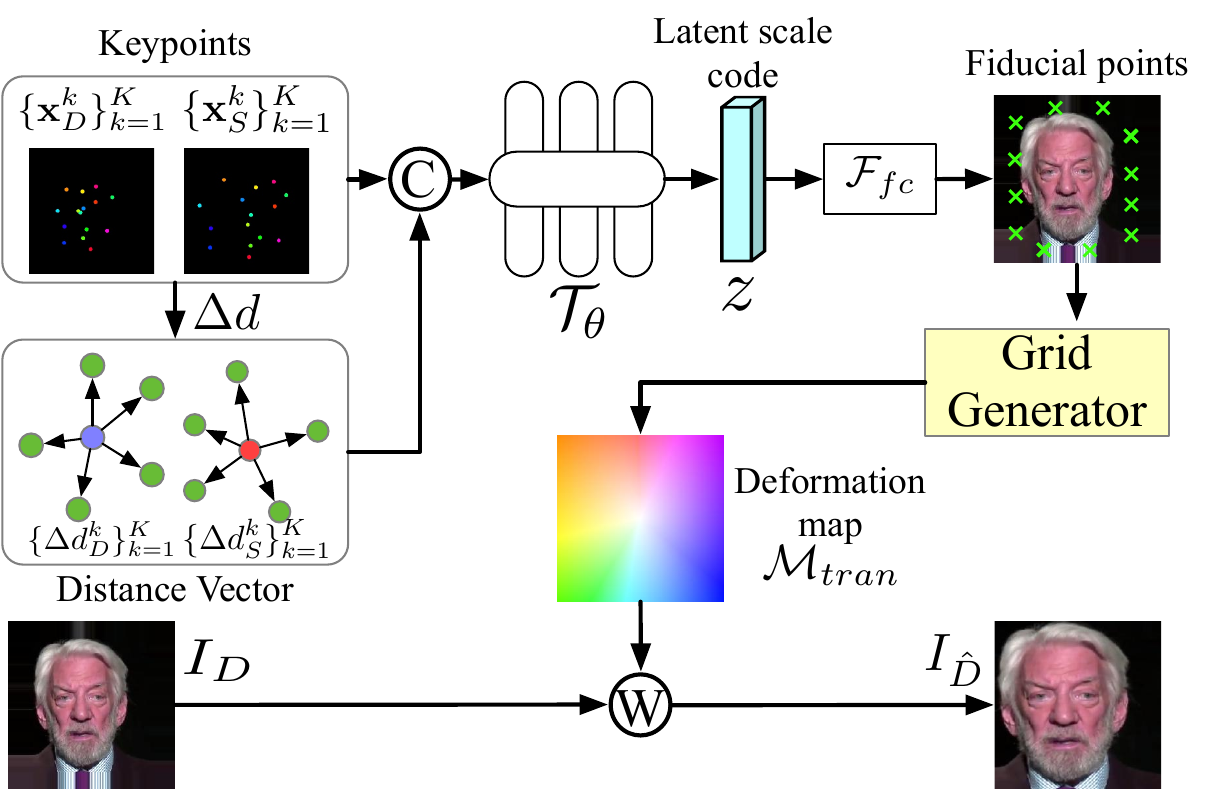}
    \vspace{-10pt}
    \caption{Illustration of the proposed scale transformation module. The symbol \textcircled{w} denotes the warping operation, and the \textcircled{c} represents the concatenation. Inspired by~\cite{shi2016robust}, we utilize the scale information maintained in the detected keypoints to predict the fiducial points. Then we feed the fiducial points into the grid generator to produce the deformation map.} 
    \vspace{-10pt}
    \label{fig:stn}    
\end{figure}

\noindent\textbf{Scale transformation.}
\label{sec:transform}
Taking the driving image $I_{D}$ of any scale as input, we aim to learn a scale transformation module (see Figure~\ref{fig:stn}) that is capable of aligning the driving facial scale to match that of the source image. Because the detected keypoints contain the scale information of the faces~\cite{zhao2022thin}, we take keypoints of the facial image detected by the keypoint detector $\mathcal{F}_{kp}$ to participate in the scale transformation step. To provide the information of scale difference, we not only feed source keypoints which indicate the source facial scale but also incorporate the driving facial keypoints into the scale transformation module. In this way, the scale transformation module can capture the scale difference. 

The keypoints on a face image are discretely distributed and their scale expression is implicit. To better extract scale information, we attempt to explore its structure information~\cite{tao2022structure} by calculating the distance vector between each keypoint and the centroid of all keypoints in the image:
\begin{equation}
    \Delta d^k_{\tau} = \mathbf{x}^k_\tau - \overline{\mathbf{x}}_\tau, \tau\in \{D, S\}, k=1,\dots K
\end{equation}
where the $\overline{\mathbf{x}}_\tau$ is the centroid of all keypoints : $\overline{\mathbf{x}}_\tau = \frac{1}{K}\sum_{k=1}^K\mathbf{x}_\tau^k $. The distance vector $\Delta d^k_{\tau}$ enables our method to realize the topological structure of the face. 

Then, we encode the keypoints and the distance vector into a latent scale code which can be utilized to represent the scale information of both images.
We design a scale-aware localization network to predict a set of fiducial points $\mathbf{C} = \{\mathbf{c}_i\}_{t=1}^T$ with the input of the keypoints by directly regressing their $x,y$-coordinates:
\begin{equation}\label{eq:fiducial}
    \begin{aligned}
        z = \mathcal{T}_\theta( \{\mathbf{x}_{S}^k\}_{k=1}^K||\{\mathbf{x}_{D}^k\}_{k=1}^K &||\{\Delta d^k_{S}\}_{k=1}^K ||\{\Delta d^k_{D}\}_{k=1}^K) \\
        % z &= \mathcal{T}_\theta(e)\\
        \{\mathbf{c}_i\}_{t=1}^T &= \mathcal{F}_{fc}(z)\\
    \end{aligned}
\end{equation}
where ``$||$'' is the concatenation operator, and $\mathcal{T}_\theta$ represents the scale-aware localization network, which is parameterized by $\theta$. We implement the $\mathcal{T}_\theta$ as a multiple-layer perceptron (MLP). The $\mathcal{F}_{fc}$ is a fully connected layer to regress the coordinates of the fiducial points. The $z$ is an intermediate latent scale code which will be utilized to aggregate the scale information into the generation process in Sec.~\ref{sec:inject}. Our scale transformation module localizes fiducial points based on the scale information. It is expected to capture the overall scale difference between the source and the driving image and localizes fiducial points accordingly. 

Immediately, we adopt the grid generator from the~\cite{shi2016robust,jaderberg2015spatial} to produce a scale deformation map $\mathcal{M}_{tran}$ with the predicted fiducial points $\{\mathbf{c}_{t=1}^T\}$ as input. Then, we can produce a scale rectified image $I_{\hat{D}}$ by warping the driving image $I_{D}$ given deformation map $\mathcal{M}_{tran}$:
\begin{equation}
    I_{\hat{D}} = \mathcal{F}_{warp}(I_{D}, \mathcal{M}_{tran})
\end{equation}
where $\mathcal{F}_{warp}(\cdot,\cdot)$ indicates the warping function. In this way, we perform scale alignment on the driving images to match the scale of the source images. 

To learn our scale transformation module, we utilise the ground truth $I_{GT}$ to constrain the rectification of the driving face. We build a reconstruction loss between the rectified image $I_{\hat{D}}$ and the ground truth $I_{GT}$ by a pre-trained VGG-19~\cite{johnson2016perceptual} network at multi-resolution:
\begin{equation}\label{eq:rec}
    \mathcal{L}_{rec}^{\hat{D}} = \sum_{i,j} |V_i(I_{GT})-V_i(I_{\hat{D}})|
\end{equation}
where $V_i$ is the $i^{th}$ layer of the VGG-19 network, and $j$ indicates that the image is downsampled $j$ times. Therefore, our scale transformation module can correctly rectify the driving facial scale with the supervision from Eq.~\ref{eq:rec}.

\begin{figure}[t]
  \centering
    \includegraphics[width=1\linewidth]{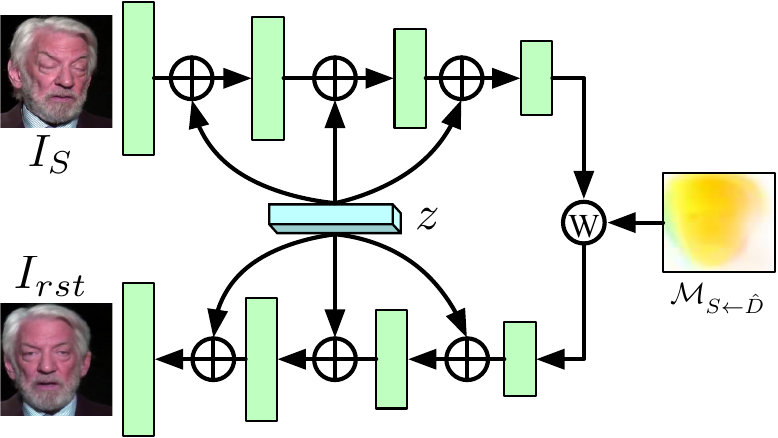}
    \vspace{-10pt}
    \caption{Illustration of the scale embedding in the generation process. The symbol \textcircled{w} denotes the warping operation, and the symbol \textcircled{+} represents the element-wise addition operator.} 
    \vspace{-10pt}
    \label{fig:inject}    
\end{figure}

\vspace{-3pt}
\subsection{Multi-layer scale information embedding} 
\vspace{-3pt}
\label{sec:inject}
Scale information can be progressively degraded during face generation as it passes through multiple layers. As a result, the facial scale can not be maintained correctly in the process of face generation. Therefore, in addition to adjusting the scale of the driving image to match the source image at the beginning, we also aggregate the scale information into the generation process to assist the synthetic face in maintaining the source facial scale. 

\noindent\textbf{Scale embedding.} The latent scale code is a crucial component in our method as it enables us to localize the fiducial points and rectify the scale of the driving face to match that of the source face. Consequently, the latent scale code potentially contains the scale information of the source face. As shown in Figure~\ref{fig:inject}, motivated by the classifier-free diffusion model~\cite{ho2022classifier}, we directly insert the latent scale code $z$ learned in the scale-ware transformation module (see Eq.~\ref{eq:fiducial}) into each layer of generation process by an additive operation. In this way, we can avoid the degradation of source scale information during generation.

We first utilize a convolutional block $\mathcal{F}_{en}$ to encode the source image $I_S$: $\mathbf{f}_1 = \mathcal{F}_{en}(I_S)$. Then we put it into the encoder-decoder generation process:
\begin{equation}
    \begin{aligned}
    \mathbf{f}_{i+1} &= \mathcal{F}_{down}^{i}(\mathbf{f}_{i}+W_{i}z), i = 1,\dots, L-1\\
    \mathbf{f}_L^w &= \mathcal{F}_{warp}(\mathbf{f}_L, \mathcal{M}_{S\leftarrow \hat{D}}) \\
    \mathbf{f}_{i}^w &= \mathcal{F}_{up}^{i}(\mathbf{f}_{i+1}^w+W_{i}^wz), i = 1,\dots, L-1\\
     \end{aligned}
\end{equation}
where $\mathcal{F}^i_{down}(\cdot),i=1,\dots, L-1$ are a set of convolutional blocks to downsample the input image, and $\mathcal{F}^i_{up}(\cdot),i=1,\dots, L-1$ are used for upsampling the input feature map. $W_{i}^w$ and $W_{i}$ are learnable parameters to change the dimention of $z$. In the middle of the generation process, we utilize the motion map $\mathcal{M}_{S\leftarrow \hat{D}}$ estimated by dense motion block~\cite{siarohin2019first} (as shown in Figure~\ref{fig:framework}) to warp the intermediate feature map. To obtain the high semantic information of the human faces, we set the $L = 4$ as the same as TPSM~\cite{zhao2022thin}.

In this way, we aggregate the latent scale code into each layer of the generation process to prevent the degradation of facial scale information. By applying the additive operation, we do not increase the complexity of the network while introducing the scale information into the generation process. Finally, we feed the $\mathbf{f}_1^w$ into a $1\times1$ convolutional layer followed by a sigmoid layer to produce the final result $I_{rst}$.
The generation process takes the latent scale code $z$ as a condition and produces a scale-guided result, which contains the accurate scale as the same as the source face.
% \begin{equation}
    % I_{rst} = \mathcal{F}_{toRBG}(\mathbf{f}_{0}^w)
% \end{equation}
% where $\mathcal{F}_{toRBG}$ consists of a $1\times1$ convalutional layer and a sigmoid layer. 
% Taking the scale information into the consideration, the final result $I_{rst}$ .

\vspace{-3pt}
\subsection{Training}\label{sec:training}
\vspace{-3pt}
\noindent\textbf{Scale removal.}
In the training stage, similar to~\cite{siarohin2019first,hong2022depth}, we sample two images from one video as the source image and the driving image, respectively. Thereby the facial scale of the two images is the same, which is not helpful for learning our network. 
% In practice, the talking head video generation task involves animating the expression and pose of the driving face. 
Therefore, to maintain the original pose and expression of the face while removing its scale, we introduce an expression-preserved augmentation to augment the scale of driving face online at the training stage.

Our expression-reserved augmentation contains two types of augmentations, including horizontal and vertical augmentation. Horizontal augmentation aims to squeeze or stretch the image horizontally while keeping its height unchanged, while vertical augmentation operates in a vertical direction. The height and width of the distorted image are $\alpha H$ and $\beta W$, respectively, where the $H$ and $W$ are the height and width of the original image and the $\alpha$ and $\beta$ the augmented factor of horizontal and vertical augmentation. We randomly choose a new $\alpha$ and $\beta$ from the range $[1-\delta, 1+\delta]$ during each forward pass. After that, we pad the image to a square shape and then resize it to the original size ($H \times W$). This way, the traces of the above two augmentations will be preserved.

Using our expression-preserved augmentation, the expression and pose are maintained in the augmented image  while the scale of the face has been changed. 

\begin{figure*}[ht]
  \centering
    \includegraphics[width=0.96\linewidth]{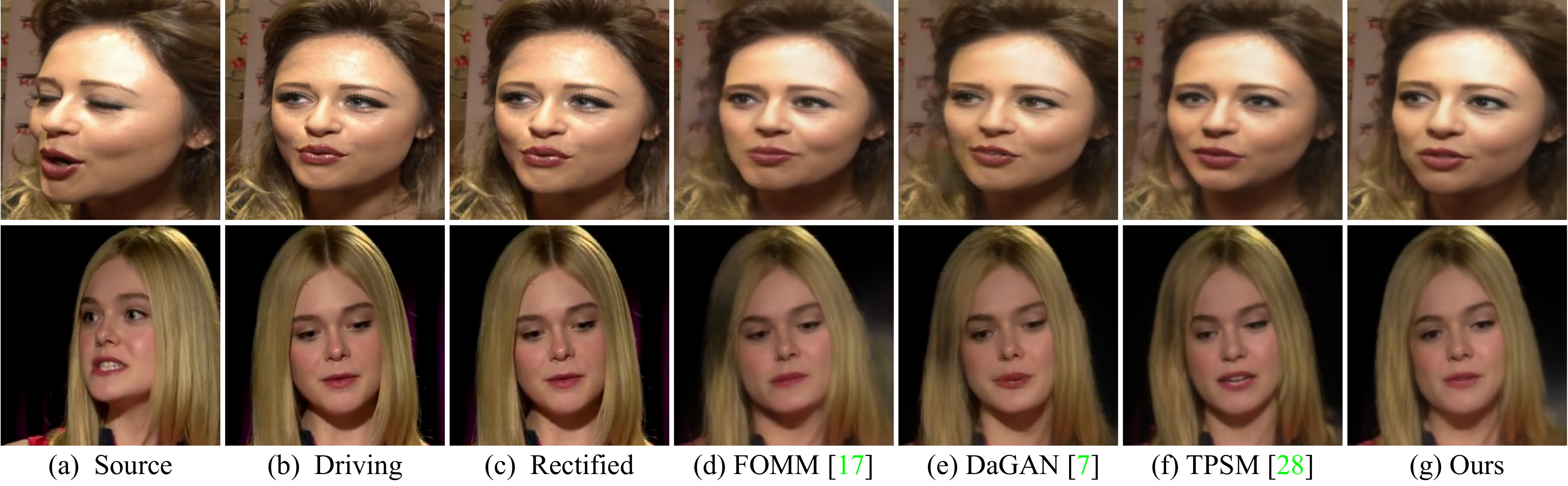}
    \vspace{-5pt}
    \caption{Qualitative comparisons of same-scale same-identity reenactment on the VoxCeleb1 dataset} 
    \vspace{-5pt}
    \label{fig:same-scale}    
\end{figure*}
\noindent\textbf{Optimization.}
Following FOMM~\cite{siarohin2019first} and DaGAN~\cite{hong2022depth}, we utilize a pre-trained VGG-19~\cite{johnson2016perceptual} network to calculate the reconstruction loss $\mathcal{L}_{rec}^{D}$ between the original driving image $I_D$ and the generated image $I_{rst}$ at multi-resolutions. 
Additionally, we also take an equivariance loss $\mathcal{L}_{eq}$~\cite{siarohin2019first} to constrain the kyepoints detection. Furthermore, we incorporate the keypoints distance loss $\mathcal{L}_{dist}$~\cite{wang2021one} to prevent the identified keypoints from clustering in a confined area. Therefore, we train our framework in an end-to-end manner using the following total objective function:
\begin{equation}
    \mathcal{L} = \lambda_1(\mathcal{L}_{rec}^D+\mathcal{L}_{rec}^{\hat{D}})+\lambda_2\mathcal{L}_{eq}+\lambda_3\mathcal{L}_{dist}
\end{equation}
Where the $\lambda_{1}$, $\lambda_{2}$, $\lambda_{3}$ are the hyper-parameters to allow for balanced learning from these losses. These losses are detailed in the \emph{Supprementary Material}.

\begin{figure*}[t]
  \centering
    \includegraphics[width=0.96\linewidth]{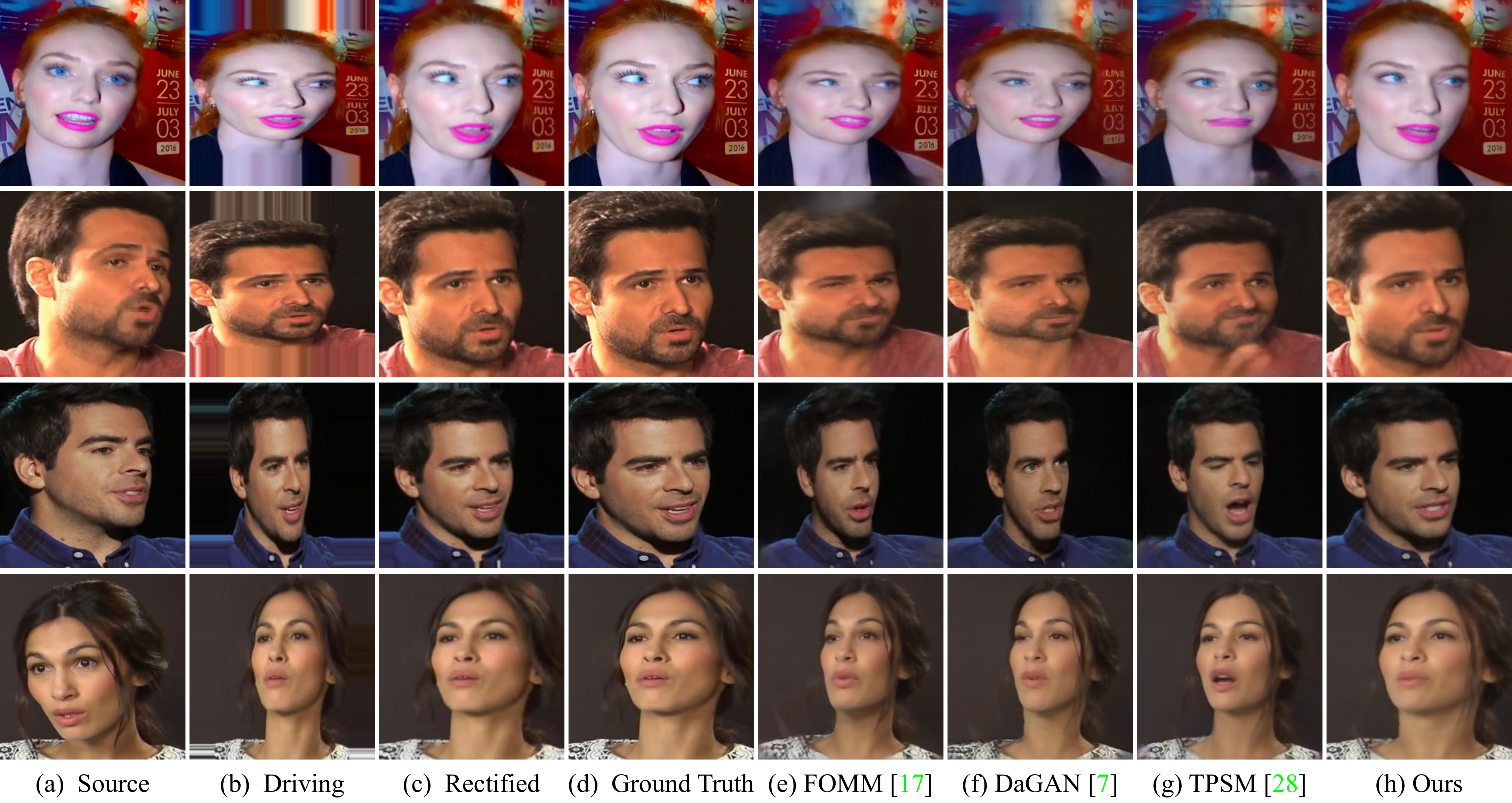}
    \vspace{-5pt}
    \caption{Qualitative comparisons of different-scale same-identity reenactment on the VoxCeleb1 dataset} 
    \vspace{-10pt}
    \label{fig:different-scale}    
\end{figure*}

\begin{figure*}[t]
  \centering
    \includegraphics[width=0.965\linewidth]{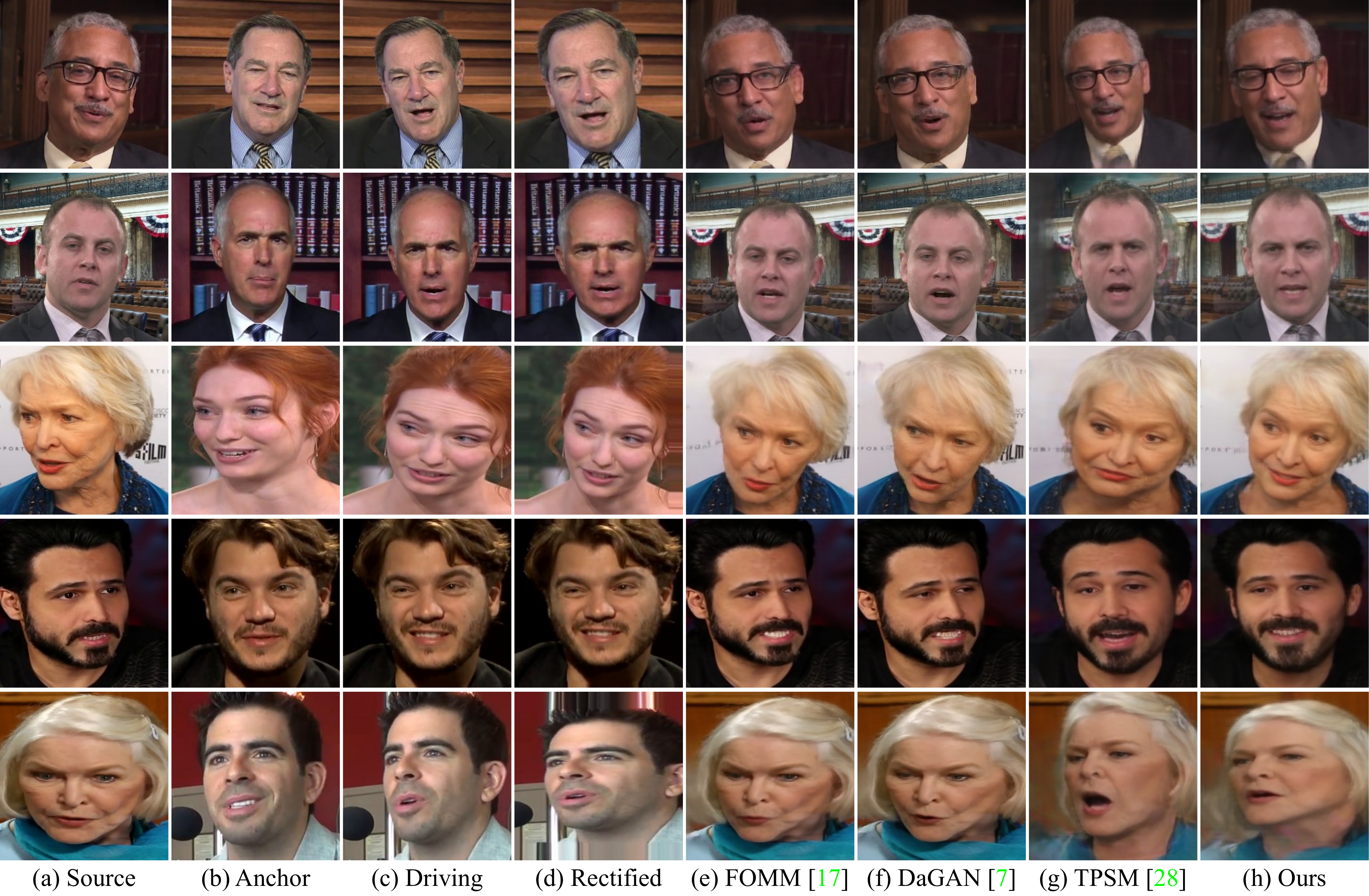}
    \vspace{-2pt}
    \caption{Qualitative comparisons of cross-identity reenactment on the VoxCeleb1~\cite{nagrani2017voxceleb} and HDTF\cite{zhang2021flow} Dataset. The first two rows are from the HDTF\cite{zhang2021flow} dataset and the last three are from the VoxCeleb1~\cite{nagrani2017voxceleb}. More samples are reported in \emph{Supplementary Material}.} 
    \vspace{-10pt}
    \label{fig:cross-id}    
\end{figure*}

\vspace{-5pt}
\section{Experiments}
\vspace{-5pt}
In this section, we perform extensive experiments on talking head video generation benchmarks to evaluate our proposed method. We also provide additional experimental results and video results in \emph{Supplementary Material}.
\vspace{-3pt}
\subsection{Dataset and Metrics.}
\vspace{-3pt}
\noindent\textbf{Dataset.} 
In this work, our experimentation focuses on talking head generation and utilizes two datasets, VoxCeleb1~\cite{nagrani2017voxceleb} and HDTF~\cite{zhang2021flow}. We employ the same sampling strategy used in DaGAN~\cite{hong2022depth} for the same-scale same-identity test set to evaluate same-identity reenactment. Additionally, we test the generalization ability of our model by applying it to cross-identity reenactment on the HDTF dataset, despite being trained on the VoxCeleb1 dataset.

\noindent\textbf{Protocal of different scale evaluation.} 
% In this work, we aim to address the scale-inconsistent problem between the source and driving face. Thus, we first create a test set in which the scales of source and driving faces are different while we still can obtain the ground truth to evaluate our method quantitatively. Taking the original testing pair as $<I_S, I_D>$ in same-scale same-identity test set, we utilize the expression-preserved augmentation to augment the driving image to produce a new testing pair $<I_S,I_{D^\prime}>$, in which the source and driving face have different scale. We collect $4166$ image pairs in this way and refer it as the different-scale same-identity test set. In this way, we can utilize the original driving image $I_D$ as the ground truth to calculate the evaluation metrics. Taking practical application into consideration, we randomly choose the $\alpha$ and $\beta$ of expression-preserved augmentation from the range $[1-\delta, 1+\delta]$ for each testing pair and set $\delta = 0.3$.
Our objective in this study is to address the problem of scale inconsistency between the source and driving faces. To achieve this, we first create a test set that includes faces of different scales while still enabling us to evaluate our method using ground truth quantitatively. Specifically, we use expression-preserved augmentation to augment the driving image and create a new testing pair $<I_S,I_{D^\prime}>$, in which the source and driving face have different scales while keeping the original testing pair $<I_S, I_D>$ in the same-scale same-identity test set. Using this method, we collect 4166 image pairs and refer to it as the different-scale same-identity test set. By using the original driving image $I_D$ as the ground truth ($I_{GT}$), we are able to calculate the evaluation metrics. To ensure that our method is applicable in practical scenarios, we randomly select the values of $\alpha$ and $\beta$ for expression-preserved augmentation from the range of $[1-\delta, 1+\delta]$ for each testing pair, where $\delta = 0.3$.

\noindent\textbf{Metrics.} In this section, we utilize several metrics to evaluate the quality of the generated images. Specifically, we adopt the structured similarity (SSIM), peak signal-to-noise ratio (PSNR), learned perceptual image patch Similarity (LPIPS)~\cite{zhang2018unreasonable}, and $\mathcal{L}_1$ distance to evaluate the low-level similarity between the generated and driving images. Additionally, we take two other metrics, Average Keypoint Distance (AKD), and Average Euclidean Distance (AED) proposed in FOMM~\cite{siarohin2019first} to evaluate the keypoint-based methods. 

% \subsection{Implementation Details.}
% The structure of the keypoint detector and dense motion module is borrowed from the FOMM~\cite{siarohin2019first}. We report the architecutre of each multiple layer perceptron (\ie $\mathcal{T}_\theta$, $\mathcal{F}_{down}^i$ and $\mathcal{F}_{up}^i$) in \emph{Supplementary Material}. For the training objective, we set the hyper-parameters $\lambda_1 = \lambda_2 = \lambda_3 = 10$. The keypoint number of this method is $15$, which is the same as DaGAN~\cite{hong2022depth}. We set the number of the fiducial points as $20$. For the expression-preserved augmentation, we set the scale factor as 0.3. Different from the MRAA~\cite{siarohin2021motion} and TPSM~\cite{zhao2022thin}, we train our method in an end-to-end manner using $8$ RTX 3090 GPUs and each GPU takes $8$ training pairs as input.
\vspace{-3pt}
\subsection{Comparison with State-of-the-art Methods.}
\vspace{-3pt}
\begin{table}[t]
  \centering
  \resizebox{1\linewidth}{!}{
        \begin{tabular}{l|cccccc}
        \toprule
        \multirow{2}*{Model} & \multicolumn{6}{c}{(a) Results of Same-identity Reenactment on VoxCeleb1}\\
        %  \midrule
         & SSIM (\%) $\uparrow$ & PSNR $\uparrow$  &  LPIPS $\downarrow$ & $\mathcal{L}_1$ $\downarrow$ & AKD $\downarrow$ &  AED $\downarrow$\\
        \midrule
        X2face~\cite{wiles2018x2face} &0.719 & 22.54  & - & 0.0780&7.687&0.405  \\
        % NeuraHead-FF(\cite{zakharov2019few}) &63.5 & 20.82  & - & - &- & - & - & - &  0.722 & 3.30\\
        marioNETte~\cite{ha2020marionette} &0.755 & 23.24 & - & - &- & - \\
      FOMM~\cite{siarohin2019first}) & 0.723 & 30.39  & 0.199  & 0.0430 & 1.294 & 0.140  \\
      MeshG~\cite{yao2020mesh}& 0.739&30.39 & - & - & - & -\\
      face-vid2vid~\cite{wang2021one}& 0.761& 30.69 & 0.212 & 0.0430 & 1.620 & 0.153 \\
      MRAA~\cite{siarohin2021motion}& 0.800  & 31.39  & 0.195 & 0.0375 & 1.296 &0.125  \\
      DaGAN~\cite{hong2022depth}& 0.804 &31.22  & 0.185  & 0.0360 &1.279 & 0.117  \\
      TPSM~\cite{zhao2022thin}&  0.816  & 31.43  & 0.179 &0.0365 & 1.233 &0.119  \\
        \midrule
        OSTNet (Ours) & \textbf{0.820} & \textbf{31.91}  & \textbf{0.176} & \textbf{0.0335} & \textbf{1.230} & \textbf{0.114} \\
        \bottomrule
        \end{tabular}
}
\caption{Comparisons with state-of-the-art methods on the same-scale same-identity reenactment on VoxCeleb1.}
\vspace{-15pt}
\label{tab:same-scale}
\end{table}

\noindent\textbf{Same-scale same-identity reenactment.} To ensure a fair comparison with other methods, we first evaluate our approach on the same-scale same-identity test set, as done in previous works~\cite{siarohin2019first,hong2022depth}. We present the results in Table~\ref{tab:same-scale} and Figure~\ref{fig:same-scale}. Despite being designed to address scale-related issues, our method achieved the best performance on the VoxCeleb1 dataset. We employ the same keypoint detector and dense motion as DaGAN~\cite{hong2022depth} and FOMM~\cite{hong2022depth}. However, our OSTNet module produces better results, with our approach achieving an SSIM metric score of $0.820$, compared to DaGAN's score of $0.804$, resulting in a $1.6\%$ improvement. This demonstrates that incorporating scale information during talking face generation can produce higher-quality results.
Furthermore, our method obtains the highest score of $0.114$ on the facial identity metric (AED), indicating that our approach better preserves the identity of the source face during generation. Notably, our method outperforms all other methods across all metrics presented in Table~\ref{tab:same-scale}. This provides strong evidence for the superiority of our approach, especially considering that it was explicitly designed for different-scale talking face generation.

\noindent\textbf{Different-scale same-identity reenactment.}
In addition to the scale-consistent reenactment experiment, we also conduct a different-scale same-identity reenactment experiment. The driving face (Figure~\ref{fig:different-scale}(b)) of the different-scale test set is obtained from the original face of the same-scale same-identity test set after applying our proposed expression-preservation augmentation. The different-scale same-identity reenactment experiment results are presented in Table~\ref{tab:different-scale} and Figure~\ref{fig:different-scale}. Our method improves significantly compared to other state-of-the-art methods from the observation of Table.~\ref{tab:different-scale}. \eg, our SSIM outperforms that of TPSM~\cite{zhao2022thin} by $18\%$. We utilize the scale transformation module to align the driving facial scale to match the source facial scale, leading to rectified faces (Figure~\ref{fig:different-scale}(c)). Compared with Figure~\ref{fig:different-scale}(c) and Figure~\ref{fig:different-scale}(d), our scale transformation can recover the original scale of the driving faces. These results presented in Table~\ref{tab:different-scale} and Figure~\ref{fig:different-scale} strongly suggest that our method can indeed rectify the scale of the driving face to match that of the source face and synthesize high-quality face image with an accurate scale. 
\begin{table}[t]

  \centering
  \resizebox{1\linewidth}{!}{
        \begin{tabular}{l|cccccc}
        \toprule
        \multirow{2}*{Model} & \multicolumn{6}{c}{(a) Results of Same-identity Reenactment on VoxCeleb1}\\
        %  \midrule
         & SSIM $\uparrow$ & PSNR $\uparrow$  &  LPIPS $\downarrow$ & $\mathcal{L}_1$ $\downarrow$ & AKD $\downarrow$ &  AED $\downarrow$\\
        \midrule
       
      FOMM~\cite{siarohin2019first}) & 0.581 & 29.39  & 0.350  & 0.0876 & 4.785 & 0.374  \\
      face-vid2vid~\cite{wang2021one}& 0.615& 29.42 & 0.320 & 0.0782 & 4.218 & 0.335 \\
      MRAA~\cite{siarohin2021motion}& 0.617  & 29.63  & 0.337 & 0.0815 & 4.617 &0.348  \\
      DaGAN~\cite{hong2022depth}& 0.605 &29.57  & 0.333  & 0.0829 &4.569 & 0.350  \\
      TPSM~\cite{zhao2022thin}& 0.600  & 29.43  & 0.341 &0.0842 &4.802 &0.354  \\
        \midrule
        OSTNet (Ours) &\textbf{0.780} & \textbf{31.25} & \textbf{0.198}& \textbf{0.0408} & \textbf{1.752} & \textbf{0.142} \\
        \bottomrule
        \end{tabular}
}
\caption{Comparisons with state-of-the-art methods on different-scale same-identity reenactment on VoxCeleb1.}
\vspace{-18pt}
\label{tab:different-scale}
\end{table}

\noindent\textbf{Cross-identity reenactment.}
Due to the absence of ground truth, we conduct qualitative evaluations on Voxceleb1~\cite{nagrani2017voxceleb} and HDTF~\cite{zhang2021flow} datasets to explore the potential for cross-identity facial motion transfer. Cross-identity facial reenactment is the critical application of the talking head generation. The input (source image and driving video) of the talking head model in practice usually has different identities. From Figure~\ref{fig:cross-id}, it can be observed that existing methods such as FOMM~\cite{siarohin2019first} and DaGAN~\cite{hong2022depth} encounter difficulty in finding a well-aligned anchor frame to execute relative motion transfer. This results in imprecise pose and expression animation as depicted in Figure~\ref{fig:cross-id}(e)(f). While TPSM~\cite{zhao2022thin} tries to adjust the distribution of keypoints to match that of the source face, their results are still impacted by the driving facial scale, as seen in Figure~\ref{fig:cross-id}(g). In contrast, our method can automatically rectify the driving face to match the source face, as shown in Figure~\ref{fig:cross-id}(d), thus enabling accurate facial expression animation (Figure~\ref{fig:cross-id}(h)).
\begin{figure}[t]
  \centering
    \includegraphics[width=1\linewidth]{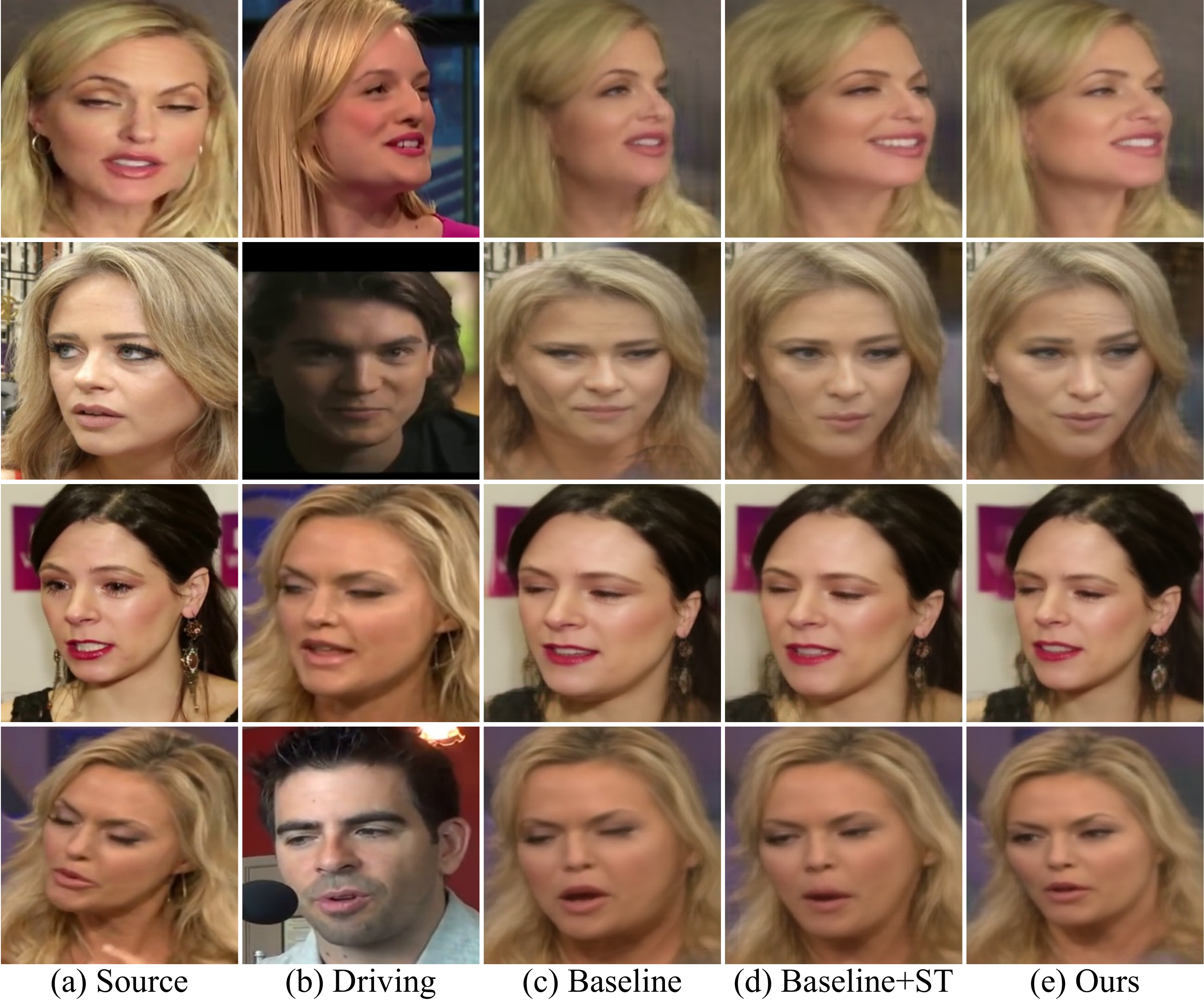}
    \vspace{-10pt}
    \caption{The illustration of Ablation study. ``Baseline'' indicates the simplest model that removes the scale transformation (ST) in Sec.~\ref{sec:transform} and scale embedding (SE) in Sec.~\ref{sec:inject} from our full method. \textbf{Zoom in for better view}.} 
    \vspace{-15pt}
    \label{fig:ablation}    
\end{figure}

\vspace{-15pt}
\subsection{Ablation study.}
\vspace{-3pt}
In this section, we perform ablation studies to qualitatively and quantitatively evaluate each component in our proposed OSTNet. The results are reported in Table.~\ref{tab:abla}, ~\ref{tab:abla-dv},~\ref{tab:generalize} and Figure~\ref{fig:ablation},. ``Baseline'' indicates the simplest model that removes the scale transformation (ST) in Sec.~\ref{sec:transform} and scale embedding (SE) in Sec.~\ref{sec:inject} from our full method.

\noindent\textbf{The effectiveness of scale transformation.} We perform the scale alignment on the driving face to match the scale of the source face. As shown in Figure~\ref{fig:different-scale}(b), the driving facial scale is augmented randomly to simulate the situation that the source face and driving face have different scales while we still can obtain the ground truth (\ie the original driving face Figure~\ref{fig:different-scale}(d)) to evaluate the result. From Figure~\ref{fig:different-scale}, we can observe that our scale transformation module can adjust the scale of the driving face to the source face so that our method does not need to find a best-aligned anchor frame in the driving video. In Figure~\ref{fig:ablation}, we find that compared with the baseline, the variance ``Baseline+ST'' eliminates the effect of driving facial scale before the generation after employing the scale transformation module. From Table.~\ref{tab:abla} and Figure~\ref{fig:ablation}, ``Baselin+ST'' can gain a great improvement compared with the ``Baseline'', which strongly verifies the effectiveness of the proposed scale transformation module.

\noindent\textbf{The effectiveness of scale embedding.}
The scale embedding is another contribution to our method. We aggregate the scale information into the generation process as introduced in Sec.~\ref{sec:inject}. As illustrated in Figure~\ref{fig:ablation} and Table.~\ref{tab:abla}, our full method ``Baseline+ST+SE'' can produce higher quality results than the ``Baseline+SE''. From Figure~\ref{fig:ablation}(d)(e), we can observe that the scale of our method with scale embedding is more accurate than that of ``Baseline+ST''. These results suggest that the scale embedding mechanism effectively enables the model to preserve the scale information of the input image during the generation process, resulting in high-quality output maintaining the source scale.
% \begin{figure}[t]
%   \centering
%     \includegraphics[width=0.7\linewidth]{figures/scale-ablation.pdf}
%     \caption{The ablation studies on the distance vector. The ``Ours w/o DV'' means we do not use the distance vector in Eq.~\ref{eq:fiducial}.} 
%     \label{fig:scale-ablation}    
% \end{figure}

\begin{table}[t]
  \centering
  \resizebox{1\linewidth}{!}{
        \begin{tabular}{lcccccc}
        \toprule
        Model & SSIM $\uparrow$ & PSNR $\uparrow$ &  LPIPS $\downarrow$ & $\mathcal{L}_1$ $\downarrow$ & AKD $\downarrow$ &  AED $\downarrow$ \\
        \midrule
        Baseline & 0.625& 29.63& 0.333 &  0.0821 & 4.546 & 0.354\\
        Baseline + ST & 0.751 & 30.76 & 0.219 & 0.0466& 2.031 & 0.161\\
        Baseline + ST + SE (OSTNet)&\textbf{0.780} & \textbf{31.25} & \textbf{0.198}& \textbf{0.0408} & \textbf{1.752} & \textbf{0.142}\\
        \bottomrule
        \end{tabular}
}
\caption{Ablation studies: ``Baseline'' indicates the simplest model without the scale transformation (ST) in Sec.~\ref{sec:transform} and scale embedding (SE) in Sec.~\ref{sec:inject}. }
\vspace{-10pt}
\label{tab:abla}
\end{table}
\begin{table}[t]
  \centering
  \resizebox{1\linewidth}{!}{
        \begin{tabular}{lcccccc}
        \toprule
        Model & SSIM $\uparrow$ & PSNR $\uparrow$ &  LPIPS $\downarrow$ & $\mathcal{L}_1$ $\downarrow$ & AKD $\downarrow$ &  AED $\downarrow$ \\
        \midrule
        Ours w/o DV & 0.816& 31.79& 0.178 &  0.0344 & 1.260 & 0.116\\
        Ours (OSTNet) & \textbf{0.820} & \textbf{31.91}  & \textbf{0.176} & \textbf{0.0335} & \textbf{1.230} & \textbf{0.114} \\
        \bottomrule
        \end{tabular}
}
\caption{The ablation studies on the distance vector on the same-scale test set. The model ``Ours w/o DV'' means we do not use the distance vector in Eq.~\ref{eq:fiducial}.}
\vspace{-10pt}
\label{tab:abla-dv}
\end{table}

\noindent\textbf{The evaluation of the distance vector in scale transformation.} We also evaluate the distance vector in Eq.~\ref{eq:fiducial}. As shown in Table~\ref{tab:abla-dv}, each metric exhibits a slight but consistent improvement compared with ``Ours w/o DV''. The distance vector ($\{\Delta d_{\tau}^k\}_{k=1}^K, \tau\in \{S, D\}$) explicitly encodes the facial structure, which aids in generating high-quality faces.

\begin{table}[t]
  \centering
  \resizebox{1\linewidth}{!}{
        \begin{tabular}{lcccccc}
        \toprule
        Model & SSIM $\uparrow$ & PSNR $\uparrow$ &  LPIPS $\downarrow$ & $\mathcal{L}_1$ $\downarrow$ & AKD $\downarrow$ &  AED $\downarrow$ \\
        \midrule
        FOMM & 0.581 & 29.39  & 0.350  & 0.0876 & 4.785 & 0.374 \\
        FOMM + ST +SE & \textbf{0.748} & \textbf{30.56} & \textbf{0.222} & \textbf{0.0481}& \textbf{1.999} & \textbf{0.163} \\
        \midrule
        DaGAN &0.605 & 29.57  & 0.333  & 0.0829 & 4.569 & 0.350  \\
        DaGAN + ST +SE & \textbf{0.768}& \textbf{30.91} & \textbf{0.208} & \textbf{0.0439}& \textbf{1.767} & \textbf{0.153} \\
        \bottomrule
        \end{tabular}
}
\caption{Generalization studies on FOMM~\cite{siarohin2019first} and DaGAN~\cite{hong2022depth}.}
\vspace{-10pt}
\label{tab:generalize}
\end{table}

\noindent\textbf{Plug-and-play experiments.} Moreover, we have integrated our method with FOMM~\cite{siarohin2019first} and DaGAN~\cite{hong2022depth} to demonstrate its flexibility in augmenting existing video generation techniques. We chose FOMM and DaGAN as robust baselines to showcase the potential of our proposed modules. The results in Table.~\ref{tab:generalize} clearly indicate a noteworthy improvement in FOMM's and DaGAN's performance when our modules are employed. These findings strongly validate the usefulness of our scale transformation approach for the task of generating talking head videos.

\vspace{-5pt}
\section{Conclusion}
\vspace{-5pt}
In this work, we propose an online scale-aligned facial reenactment network for talking head video generation (OSTNet). OSTNet learns a scale transformation module to align the scale of the driving faces to the source face so that our method can generate high-quality video with an accurate scale. Furthermore, we incorporate the scale information learned from the scale transformation module into the generation process to improve the maintenance of the source facial scale in the final results. Extensive experimental results demonstrate that our OSTNet can correctly rectify the scale of the driving face to match that of the source face online. Ablation studies clearly show that scale transformation between the source face and the driving face can benefit the talking head video generation. Our OSTNet also produces more realistic and natural-looking results compared to the state-of-the-art.

\appendix
% \onecolumn
\section*{\LARGE{Appendix}}\label{s:appendix}
\input{append}

{\small
\bibliographystyle{ieee_fullname}
\bibliography{egbib}
}

\end{document}

%% file: append.tex
We provide additional details of several aspects of the submission, including the implementation, the optimization, and the experiment results.

\section{More Implementation Details}
The structure of the keypoint detector and dense motion module is borrowed from the FOMM~\cite{siarohin2019first}. We report the architecutre of each multiple layer perceptron (\ie $\mathcal{T}_\theta$, $\mathcal{F}_{down}^i$ and $\mathcal{F}_{up}^i$) in this {Supplementary Material}. For the training objective, we set the hyper-parameters $\lambda_1 = \lambda_2 = \lambda_3 = 10$. The number of the keypoints of this method is set to $15$, being the same as DaGAN~\cite{hong2022depth}. We set the number of the fiducial points as $20$. For the data augmentation, we set the scaling factor as 0.3. We train our method in an end-to-end manner using $8$ RTX 3090 GPUs and each GPU takes $8$ training pairs as input.
% \section{More details of architecture}
% An additional depiction of our proposed data augmentation is shown in Figure.~\ref{fig:augmentation}. As illustrated in Figure.~\ref{fig:augmentation}. the data augmentation consists of two types of augmentation, including horizontal and vertical augmentation. 
% We also append an edge padding at the end and then resize to the size of $256 \times 256$.
\begin{figure*}[t]
  \centering
    \includegraphics[width=0.9\linewidth]{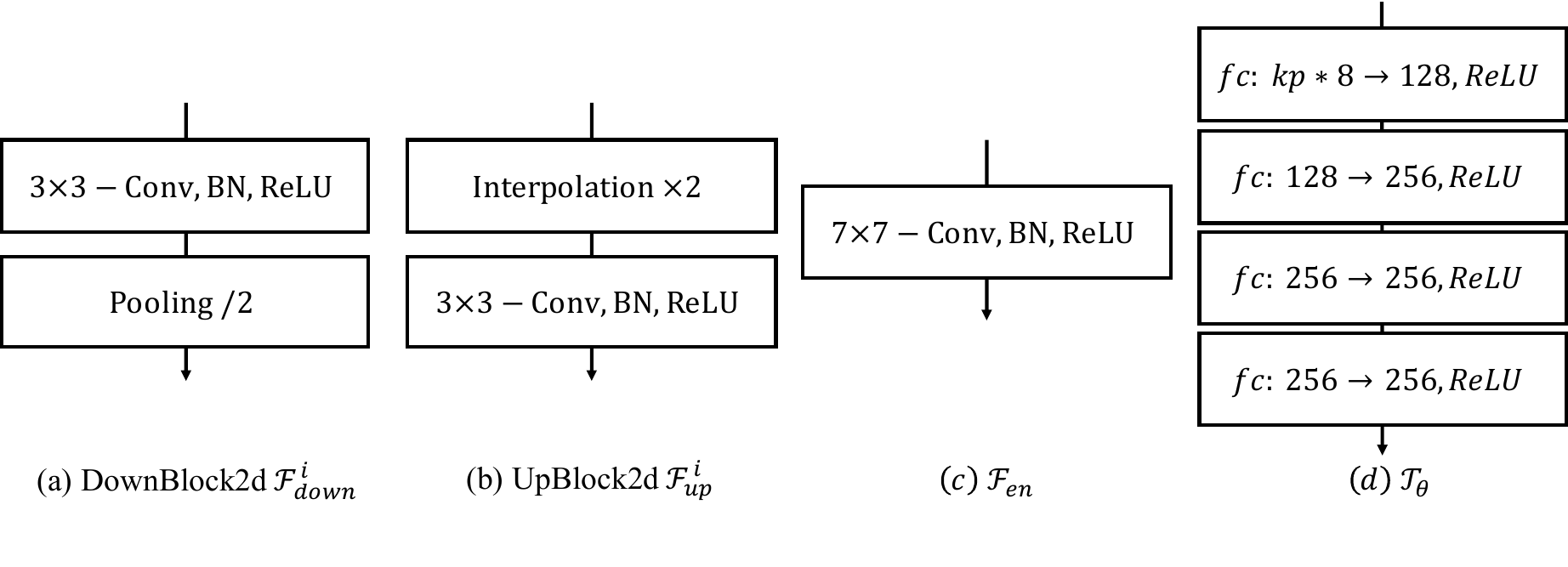}
    \vspace{-15pt}
    \caption{Illustration of detailed network structures of four different components utilized in the main framework, \ie~DownBlock2d, UpBlock2d, the image encoder, and the scale-aware localization network, in our generation framework. The “DownBlock2d” (Figure~\ref{fig:archi}a) contains a convolutional layer with a $3 \times 3$ kernel, a batch normalization layer, a ReLU activation layer, and an average pooling layer that downsamples the input. The interpolation layer in “UpBlock2d” (Figure~\ref{fig:archi}b) is utilized to upsample the image. The scale-aware localization network $\mathcal{T}_{\theta}$ consists of four fully connection layers.}
    \label{fig:archi}    
\end{figure*}
% \section{Architecture Details.}
We also depict the architecture of each sub-network of our framework in Figure~\ref{fig:archi}. We use convolutional layers with $3\times 3$ kernels in both the downsampling blocks $\mathcal{F}_{down}^i$ and the upsampling blocks $\mathcal{F}_{up}^i$. We implement the scale-aware localization network as a $4$-layer MLP as shown in Figure~\ref{fig:archi}d.

\section{More Optimization Details}
In the training stage, we take the original driving image as the ground truth $I_{GT}$.
Following FOMM~\cite{siarohin2019first} and DaGAN~\cite{hong2022depth}, we utilize a pre-trained VGG-19~\cite{johnson2016perceptual} network to calculate the reconstruction loss between the ground truth image $I_{GT}$ and the generated image $I_{rst}$ at multi-resolutions. To learn the scale transformation module, we also constraint the rectified image $I_{\hat{D}}$ to be as close as possible to the ground truth image $I_{GT}$ as follows:
\begin{equation}
    \mathcal{L}_{rec} = \sum_{l,i}(|V_{i}^l(I_{GT})-V_{i}^l(I_{rst})| + |V_{i}^l(I_{GT})-V_{i}^l(I_{\hat{D}})|),
\end{equation}
where $V_i$ is the $i^{th}$ layer of the pre-trained VGG-19 network, and $l$ indicates that the image is downsampled by $l$ times. Additionally, we also adopt an equivariance loss~\cite{siarohin2019first} to contraint the kyepoint detection: 
\begin{equation}
    \mathcal{L}_{eq} = |\mathcal{F}_{kp}(\mathcal{T}_{random}(I_S)) - \mathcal{T}_{random}(\mathcal{F}_{kp}(I_S))|,
\end{equation}
where $\mathcal{T}_{random}$ is the random nolinear transformation. In this work, we apply the random TPS transformation similar to FOMM~\cite{siarohin2019first}. Furthermore, we incorporate the keypoints distance loss $\mathcal{L}_{dist}$~\cite{wang2021one} to prevent the identified keypoints from clustering in a confined area:
\begin{equation}
    \mathcal{L}_{dist} = \sum_{i=1}^K\sum_{j=1}^K \max(0,|\mathbf{x}^i_S-\mathbf{x}^j_S|_1-\gamma), i\neq j,
\end{equation}
where $\gamma$ is a hyper-parameter set to 0.1 in this work. We train our proposed framework in an end-to-end manner using the following total objective function:
\begin{equation}
    \mathcal{L} = \lambda_1\mathcal{L}_{rec}+\lambda_2\mathcal{L}_{eq}+\lambda_3\mathcal{L}_{dist}
\end{equation}
Where $\lambda_{1}$, $\lambda_{2}$, $\lambda_{3}$ are the hyper-parameters to allow for balanced learning from these losses.

\begin{table}[!h]
  \centering
  \resizebox{1\columnwidth}{!}{
        \begin{tabular}{lcccccc}
        \toprule
        Model & SSIM $\uparrow$ & PSNR $\uparrow$ &  LPIPS $\downarrow$ & $\mathcal{L}_1$ $\downarrow$ & AKD $\downarrow$  \\
        \midrule
        FOMM~\cite{siarohin2019first} & 0.769 & 31.87 & 0.155 & 0.0363 & 1.116 \\
        face-vid2vid~\cite{wang2021one}& 0.779& 31.67 & 0.159 & 0.0366 & 1.329 \\
        
        MRAA~\cite{siarohin2021motion} & 0.794 & 32.32 & 0.156 & 0.0331 & 1.039 \\
        DaGAN~\cite{hong2022depth} & 0.823 & 32.29 & 0.136 & 0.0304 & 1.020 & \\
        TPSM~\cite{zhao2022thin} & \underline{0.860} & \underline{32.85} & \underline{0.114} & \underline{0.0264} & \underline{1.015}  \\
         \midrule
       
        Ours (OSTNet) & \textbf{0.861} & \textbf{33.54} & \textbf{0.111}& \textbf{0.0243} & \textbf{0.992}  \\
        \bottomrule
        \end{tabular}
}
\vspace{3pt}
\caption{Comparisons with state-of-the-art methods on the same-scale and the same-identity reenactment on the HDTF dataset.}
\label{tab:hdtf}
\end{table}

\begin{figure*}[t]
  \centering
    \includegraphics[width=1\linewidth]{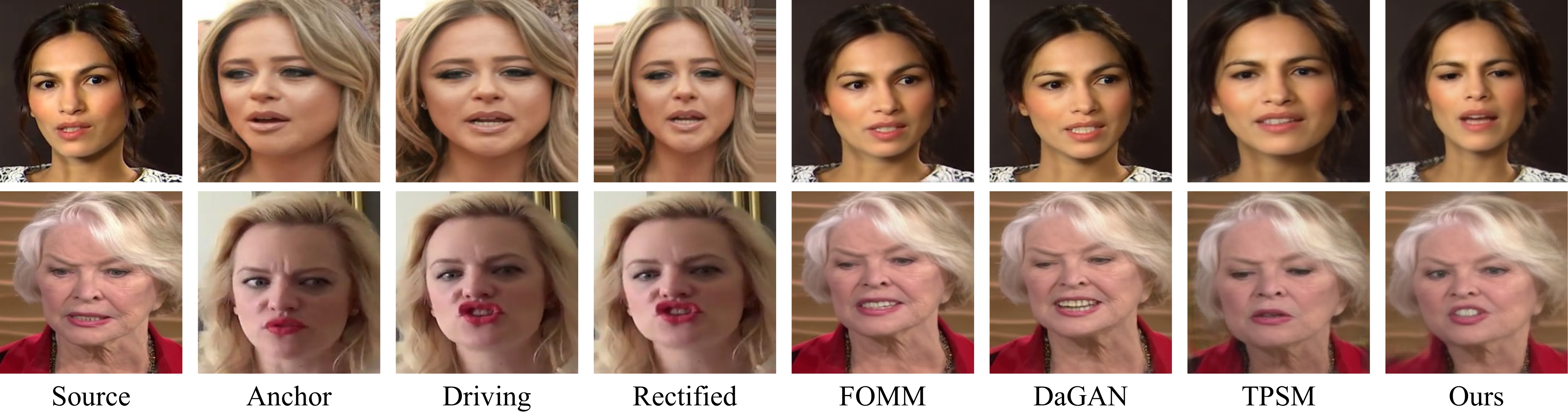}
    \caption{Qualitative comparisons of the cross-identity reenactment on  the VoxCelebv1~\cite{nagrani2017voxceleb} Dataset.}
    \label{fig:supp-vox-cross-id2}    
\end{figure*}
\section{More Experimental Results}
% \subsection{Same-scale same-identity in HDTF dataset.}
\par\noindent\textbf{Same-scale and same-identity experiment on HDTF dataset.} In this work, we also conduct the same-scale same-identity reenactment experiments on HDTF to compare with other methods. The results are reported in Table~\ref{tab:hdtf}. We can observe that our method can still obtain the best results in terms of most metrics. As we utilize the checkpoint trained on VoxCelebv1~\cite{nagrani2017voxceleb} to test on the HDTF~\cite{zhang2021flow} directly, it can also verify the generalization ability of our method.
% \subsection{More visualization on cross-identity reenactment.}
\par
\begin{figure*}[t]
  \centering
    \includegraphics[width=1\linewidth]{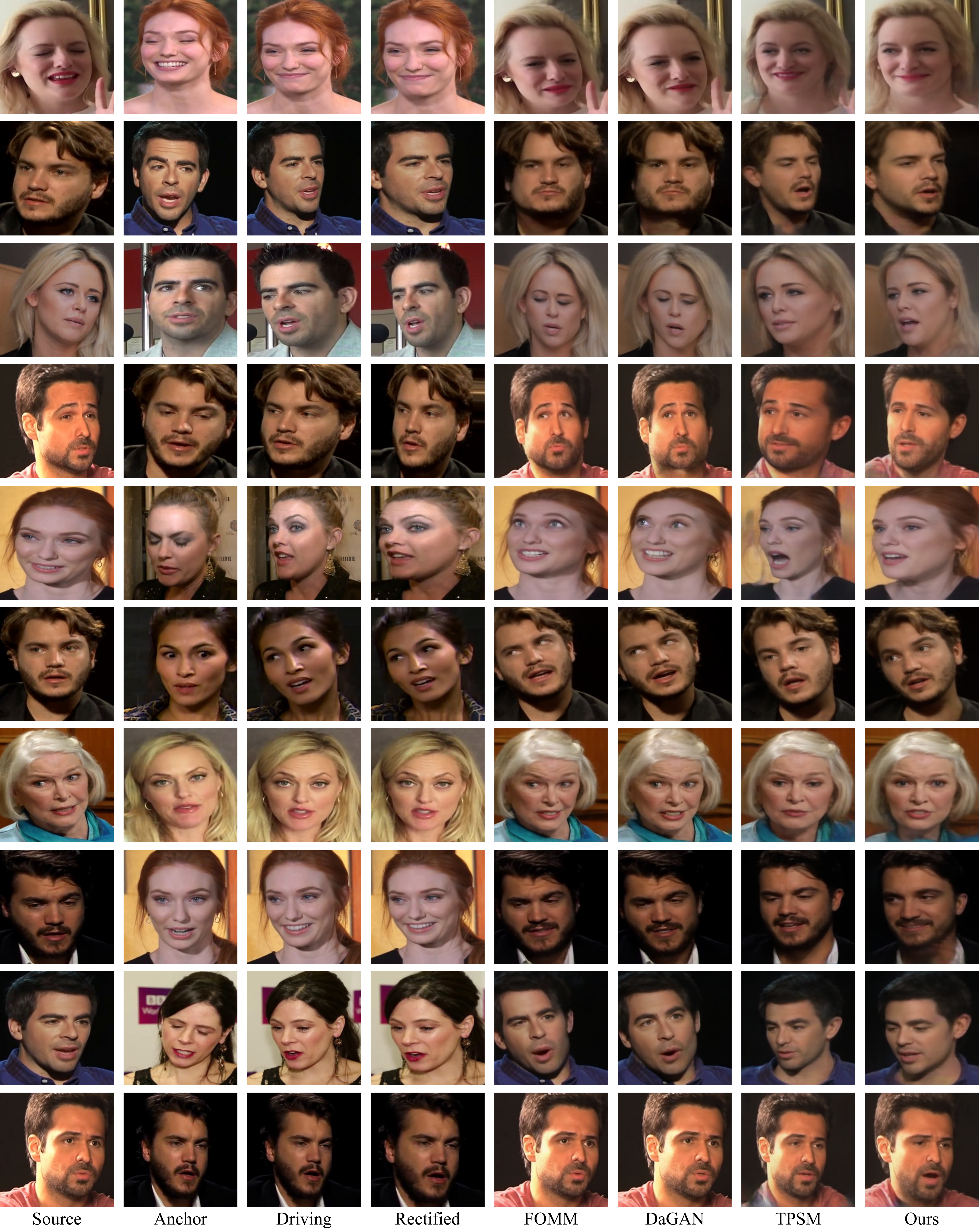}
    \caption{Qualitative comparisons of the cross-identity reenactment on  the VoxCelebv1~\cite{nagrani2017voxceleb} Dataset.}
    \label{fig:supp-vox-cross-id}    
\end{figure*}

\begin{figure*}[t]
  \centering
    \includegraphics[width=1\linewidth]{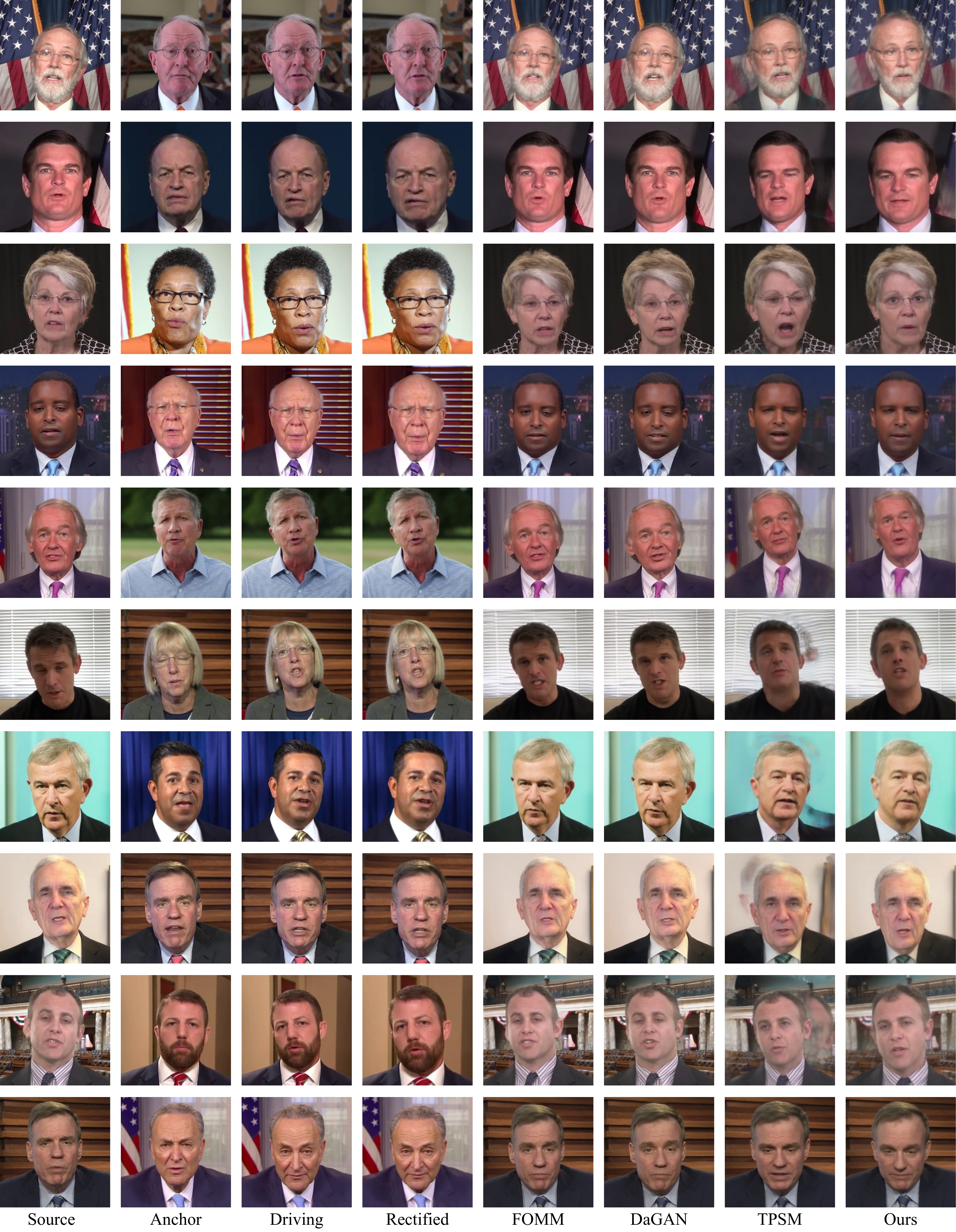}
    \caption{Qualitative comparisons of the cross-identity reenactment on the HDTF~\cite{zhang2021flow} Dataset.}
    \label{fig:supp-hdtf-cross-id}    
\end{figure*}

\noindent\textbf{More qualitative results on cross-identity reenactment.} To effectively verify the superiority of our method, we show more qualitative samples on the cross-identity reenactment experiment in Figure~\ref{fig:supp-vox-cross-id2}, Figure~\ref{fig:supp-vox-cross-id} and Figure~\ref{fig:supp-hdtf-cross-id}. These samples validate that our proposed method can indeed produce high-quality results using the proposed online scale transformation strategy.
% \subsection{Video Sample.}
\par\noindent\textbf{Video demo.} We additionally provide a video demo in the supplementary material to better show the qualitative video generation results produced by our method.

